\documentclass[11pt,a4paper]{article}
\usepackage[hyperref]{acl2021}
\usepackage{times,latexsym}

\usepackage{url}
\usepackage[T5,T1]{fontenc}
\usepackage[utf8]{inputenc}
\usepackage{booktabs}
\usepackage{multirow}
\usepackage{multicol}
\usepackage{diagbox}
\usepackage{stfloats}

\DeclareTextSymbolDefault{\ohorn}{T5}
\DeclareTextSymbolDefault{\uhorn}{T5}

\usepackage{microtype}
\usepackage{graphicx}
\usepackage{subfigure}
\usepackage{booktabs} %
\usepackage{mathtools}
\usepackage{tikz}
\usetikzlibrary{bayesnet}
\usetikzlibrary{arrows}

\usepackage{tabularx}
\usepackage{tabulary}
\usepackage{colortbl}

\usepackage{amsmath}
\usepackage{amsfonts}
\usepackage{amssymb}
\usepackage{cuted}
\usepackage{dsfont}
\usepackage{soul}

\usepackage{algorithm}
\usepackage[noend]{algorithmic}
\usepackage{etoolbox}
\usepackage{xparse}

\usepackage[noabbrev,capitalise]{cleveref}
\newcolumntype{Y}{>{\centering\arraybackslash}X}

\usepackage{enumitem}
\newlist{todolist}{itemize}{2}
\setlist[todolist]{label=$\square$}
\usepackage{pifont}

\usepackage[disable]{todonotes}
\makeatletter
\newcommand*\iftodonotes{\if@todonotes@disabled\expandafter\@secondoftwo\else\expandafter\@firstoftwo\fi}  %
\makeatother

\newcommand{\note}[4][]{\todo[author=#2,color=#3,size=\scriptsize,fancyline,caption={},#1]{#4}} %

\definecolor{olive}{HTML}{3D9970}

\newcommand{\ignore}[2][]{}

\newcommand{\rahul}[2][]{\note[#1]{RA}{yellow!20}{#2}}

\aclfinalcopy %
\def\aclpaperid{931} %

\title{Minimax and Neyman--Pearson Meta-Learning for Outlier Languages}

\makeatletter
\newcommand{\printfnsymbol}[1]{%
  \textsuperscript{\@fnsymbol{#1}}%
}
\makeatother

\author{Edoardo Maria Ponti$^{1, 2}$\thanks{~~Equal contribution}~~~~Rahul Aralikatte$^{3}$\printfnsymbol{1} \\ \textbf{Disha
Shrivastava}$^{1, 4}$~~~~\textbf{Siva Reddy}$^{1, 2}$~~~~\textbf{Anders Søgaard}$^{3}$ \\   
    $^1$Mila -- Quebec Artificial Intelligence Institute~~~$^2$McGill University\\
    $^3$University of Copenhagen~~~$^4$University of Montreal \\
    $^1${\tt \{edoardo-maria.ponti,siva.reddy,disha.shrivastava\}@mila.quebec} \\
    $^3${\tt \{rahul,soegaard\}@di.ku.dk} \\
}

\date{}

\begin{document}

\newcommand{\vtheta}{{\boldsymbol \vartheta}}
\newcommand{\vphi}{{\boldsymbol \varphi}}
\newcommand{\vups}{{\boldsymbol \upsilon}}
\newcommand{\vlambda}{{\boldsymbol \lambda}}
\newcommand{\vepsilon}{{\boldsymbol \epsilon}}
\newcommand{\calN}{{\cal N}}
\newcommand{\calO}{{\cal O}}
\newcommand{\xx}{\mathbf{x}}
\newcommand{\KL}{\text{KL}}
\newcommand{\vmu}{{\boldsymbol \mu}}
\newcommand{\vsigma}{{\boldsymbol \sigma}}
\newcommand{\vSigma}{{\boldsymbol \Sigma}}
\newcommand{\defn}[1]{{#1}}
\newcommand{\qlambda}{q_{\vlambda}}
\newcommand{\bert}{\textsc{bert}}
\newcommand{\softmax}{\mathrm{softmax}}
\newcommand{\R}{\mathbb{R}}
\newcommand{\Dtrain}{\mathcal{D}_{train}}
\newcommand{\Dval}{\mathcal{D}_{val}}
\newcommand{\D}{\mathrm{D}}

\DeclarePairedDelimiterX{\infdivx}[2]{(}{)}{%
  #1\;\delimsize\|\;#2%
}
\newcommand{\infdiv}{\mathbb{KL}\infdivx}
\newcommand{\lv}[1]{{\boldsymbol #1}}
\newcommand{\KLD}[2]{\mathbb{KL} \left( #1 \left|\right| #2 \right) }

\maketitle

\begin{abstract}

    Model-agnostic meta-learning (MAML) has been recently put forth as a strategy to learn resource-poor languages in a sample-efficient fashion. Nevertheless,
    the properties of these languages are often \textit{not} well represented by 
    those available during training. Hence, we argue that the {\em i.i.d.}\ assumption ingrained in MAML
    makes it ill-suited for cross-lingual NLP.
    In fact, under a decision-theoretic framework, MAML can be interpreted as minimising the expected risk across training languages (with a uniform prior), which is known as Bayes criterion. To increase its robustness to outlier languages,
    we create two variants of MAML based on alternative criteria: Minimax MAML reduces the {\em maximum} risk across languages, while Neyman--Pearson MAML \textit{constrains} the risk in each language to a maximum threshold.
    Both criteria constitute fully differentiable two-player games. In light of this, we propose a new adaptive optimiser solving for a local approximation to their Nash equilibrium.
    We evaluate both model variants on two popular NLP tasks, part-of-speech tagging 
    and question answering. 
    We report gains for their average and minimum performance across low-resource languages in zero- and few-shot settings, compared to joint multi-source transfer and vanilla MAML. The code for our experiments is available at \url{https://github.com/rahular/robust-maml}.
\end{abstract}

\section{Introduction}
Knowledge transfer is ubiquitous in machine learning because of the general scarcity of annotated data \citep[\textit{inter alia}]{pratt1993discriminability,caruana1997multitask,ruder2019neural}.  A prominent example thereof is transfer from resource-rich languages to resource-poor languages \citep{wu2019beto,ponti-etal-2019-towards,Ruder:2019jair}. 
Recently, Model-Agnostic Meta-Learning \citep[MAML;][]{finn2017model} has come to the fore as a promising paradigm: it explicitly trains neural models that adapt to new languages quickly by extrapolating from just a few annotated data points \cite{gu-etal-2018-meta,nooralahzadeh-etal-2020-zero,wu2020enhanced,li-etal-2020-learn}.

MAML usually rests on the simplifying assumption that the source `tasks' and the target `tasks' are 
independent and identically distributed (henceforth, \textit{i.i.d.}). However, in practice most scenarios of cross-lingual transfer violate this assumption: 
training languages documented in mainstream datasets do not reflect the cross-lingual variation, as they belong to a clique of few families, geographical areas, and typological features \cite{bender-2009-linguistically,joshi-etal-2020-state}. Therefore, the majority of the world's languages lies outside of such a clique.
As training and evaluation languages differ in their joint distribution, they are not exchangeable \citep[ch.\ 6]{ponti2021inductive,orbanz2012lecture}. Therefore, there is no formal guarantee that MAML generalises to the very languages whose need for transfer is most critical. %

In this work, we interpret meta-learning within a decision-theoretic framework \citep{bickel2015mathematical}. MAML, we show, minimises the expected risk across languages found in the training distribution. Hence, it follows a so-called Bayes criterion. What if, instead, we formulated alternative criteria geared towards outlier languages? The first criterion we propose, Minimax MAML, is designed to be robust to worst-case-scenario out-of-distribution transfer: it minimises the \textit{maximum} risk by learning an adversarial language distribution. The second criterion, Neyman--Pearson MAML, upper-bounds the risk for an arbitrary subset of languages via Lagrange multipliers, such that it does not exceed a predetermined threshold.

Crucially, both of these alternative criteria constitute competitive games between two players: one minimising the loss with respect to the neural parameters, the other maximising it with respect to the language distribution (Minimax MAML) or Lagrange multipliers (Neyman--Pearson MAML). Since an absolute Nash equilibrium may not exist for non-convex functions \citep{jin2020local}, such as neural networks, a common solution is to approximate local equilibria instead \citep{schafer2019competitive}. Therefore, we build on previously proposed optimisers  \citep{balduzzi2018mechanics,letcher2019differentiable,gemp2018global} where players follow non-trivial strategies that take into account the opponent's predicted moves. In particular, we enhance them with first-order momentum and adaptive learning rate and apply them on our newly proposed criteria.

\begin{figure}[t]
    \centering
    \includegraphics[width=\columnwidth]{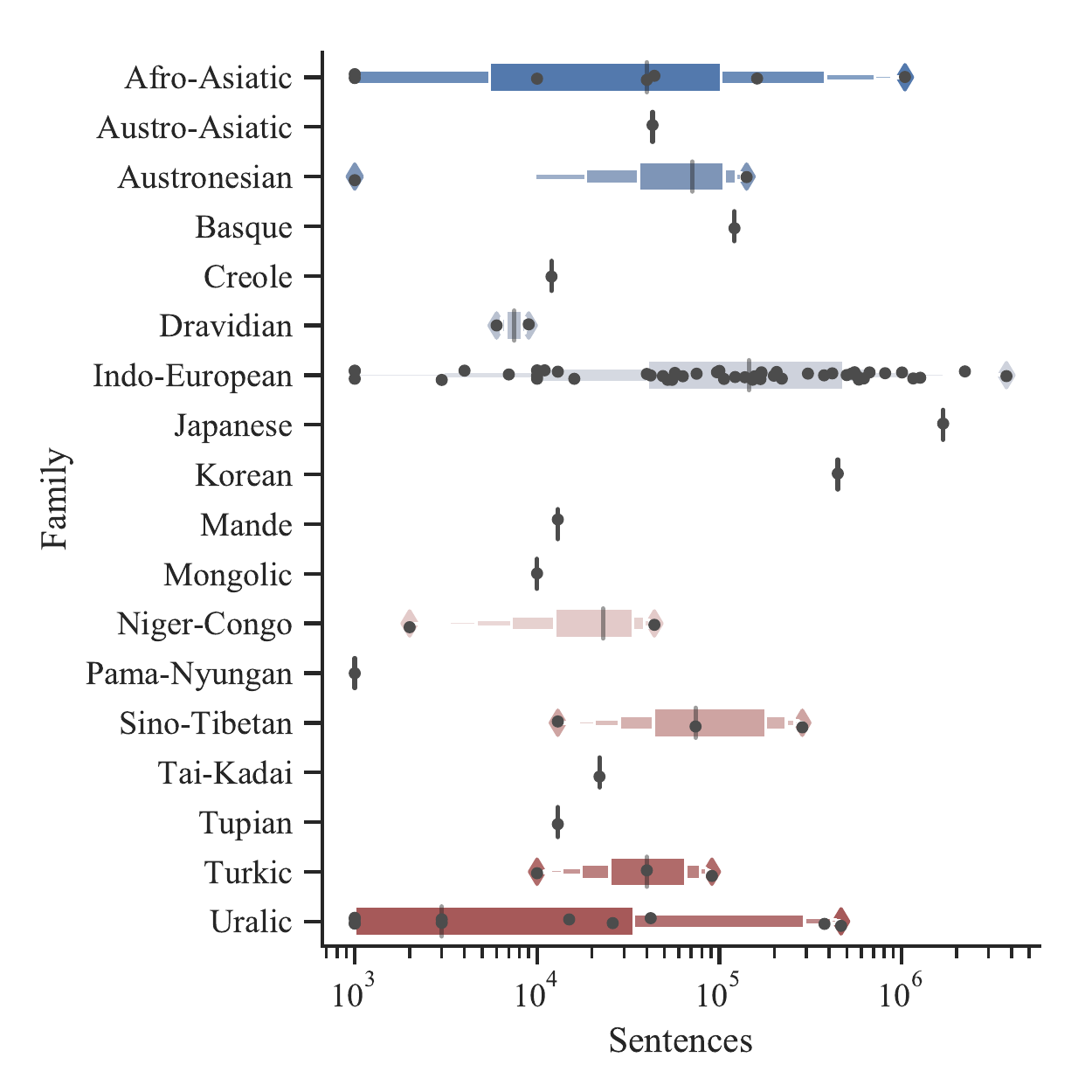}
    \caption{Annotated examples per family in the Universal Dependencies treebanks. Dots indicate individual languages, whereas boxes and whiskers mark quartiles.}
    \label{fig:ud}
    \vspace{-0.4cm}
\end{figure}

\noindent We run experiments on Universal Dependencies \citep{11234/1-3226} for part-of-speech (POS) tagging and TyDiQA \citep{tydiqa} for question answering (QA). We perform knowledge transfer to 14 and 8 target languages, respectively, which belong to under-represented and often endangered families (such as Tupian from Southern America and Pama--Nyugan from Australia). We report modest but consistent gains for the average performance across languages in few-shot and zero-shot learning settings and mixed results for the minimum performance. In particular, Minimax and Neyman--Pearson MAML often surpass vanilla MAML and multi-source transfer baselines, which are currently considered state-of-the-art in these tasks \citep{wu2019beto,ponti2020parameter,tydiqa}.

\section{Skewed Language Distributions}
\label{sec:skewlangdist}
Cross-lingual learning aims at transferring knowledge from resource-rich languages to resource-poor languages, to compensate for their deficiency of annotated data \citep{tiedemann-2015-cross,Ruder:2019jair,Ponti:2019cl}. The set of target languages ideally encompasses most of the world's languages.
However, the source languages available for training are often concentrated around few families, geographic areas, and typological features \cite{cotterell-eisner-2017-probabilistic,gerz-etal-2018-relation,gerz2018language,ponti2020xcopa,tydiqa}. As a consequence of this discrepancy, a language drawn at random might have no related languages available for training. Even when this is not the case, %
they might provide a scarce amount of examples for supervision.

\begin{figure}[t]
    \centering
    \includegraphics[width=\columnwidth]{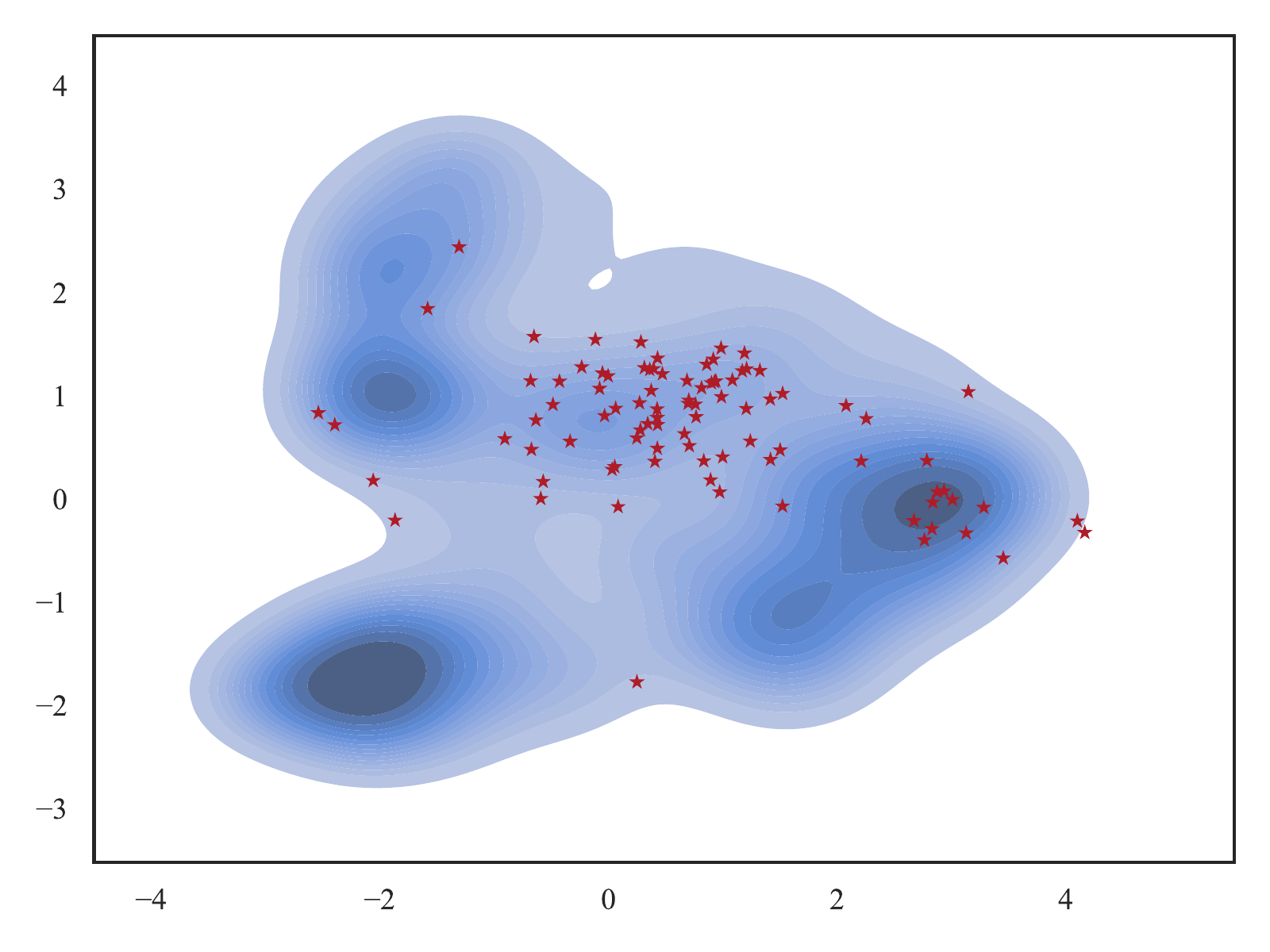}
    \caption{Density of WALS typological features of the world's languages reduced to 2 dimensions via PCA. Red dots are languages covered by UD. Darkness corresponds to more probable regions.}
    \label{fig:typology}
\end{figure}

To illustrate this point, consider Universal Dependencies \citep[UD;][]{11234/1-3226}, hitherto the most comprehensive collection of manually curated multilingual data. First, out of 245 families attested in the world according to Glottolog \citep{glottolog}, UD covers only 18.\footnote{For more details on family distributions, cf.\ \cref{fig:familydist} in the Appendix.} In fact, some families are chronically over-represented (e.g.\ Indo-European and Uralic) and others are neglected (e.g. Pama-Nyugan and Uto-Aztecan). Second, as shown in %
\cref{fig:ud}, the allocation of labelled examples across families is imbalanced (e.g.\ note the low counts for Niger--Congo or Dravidian languages). Third, one can measure how representative the linguistic traits of training languages are in comparison to those encountered around the globe. In \cref{fig:typology}, we represent UD languages as dots in the space of possible typological features in WALS \citep{wals}. These are plotted against the density of the distribution based on all languages in existence. Crucially, it emerges that UD languages mostly lie in a low-density region. Therefore, they hardly reflect the variety of possible combinations of typological features.

In general, this demonstrates that the distribution of training languages in existing NLP datasets is heavily skewed compared to the real-world distribution. Indeed, this very argument holds true \textit{a fortiori} in smaller, less diverse datasets. While this fact is undisputed in the literature, its consequences for modelling, which we expound in the next section, are often under-estimated.

\section{Robust MAML}
\label{sec:robustmaml}
Model-Agnostic Meta Learning \citep[MAML;][]{finn2017model} has recently emerged as an effective approach to cross-lingual transfer \citep{gu-etal-2018-meta,nooralahzadeh-etal-2020-zero,wu2020enhanced,li-etal-2020-learn}. MAML seeks a good initialisation point for neural weights in order to adapt them to new languages with only a few examples. To this end, for each language $\mathcal{T}_i$ a neural model $f_\vtheta$ is updated according to the loss on a batch of examples $\mathcal{L}_{\mathcal{T}_i}(f_\vtheta, \mathcal{D}_{train})$. This inner loop is iterated for \textit{k} steps.
Afterwards, the loss incurred by the model on a held-out batch $\mathcal{D}_{val}$ is compounded with those of the other languages as part of an outer loop, as shown in \cref{eq:metaobj}:
\begin{equation}\label{eq:metaobj}
\begin{split}
\vtheta^\star = \min_\vtheta \sum_{\mathcal{T}_i} \mathcal{L}_{\mathcal{T}_i} (f_{\lv\varphi_i}, \mathcal{D}_{val}) \, p(\mathcal{T}_i) \\
\text{where} \quad
\lv\varphi_i = \vtheta - \eta \nabla_\vtheta \mathcal{L}_{\mathcal{T}_i}(f_\vtheta, \mathcal{D}_{train})
\end{split}
\end{equation}
\noindent
where $\eta \in \mathbb{R}_{>0}$ is the learning rate.
Language probabilities are often taken to follow a discrete uniform distribution $p(\mathcal{T}_i) = \frac{1}{|\mathcal{T}|}$. In this case, \cref{eq:metaobj} becomes a simple average.

MAML can also be interpreted as point estimate inference in a hierarchical Bayesian graphical model (see \cref{fig:hierbayes2} in the Appendix). In this case, the adapted parameters 
$\lv\varphi_i$ are equivalent to an intermediate language-specific variable acting as a bridge between the language-agnostic parameters $\vtheta$ and the data \citep{grant2018recasting, NEURIPS2018_8e2c381d, NEURIPS2018_e1021d43}. This allows us to reason about the conditions under which a model is expected to generalise to new languages. Crucially, generalisation rests on the assumption of independence and identical distribution among the examples (including both train and evaluation), which is known as exchangeability \citep{zabell2005symmetry}. However, as seen in \cref{sec:skewlangdist},  most of the world's languages are outliers with respect to the training language distribution. Therefore, there is no solid guarantee that meta-learning may fulfil its purpose, i.e.\ generalise to \textit{held-out} languages. %

\subsection{Decision-Theoretic Perspective}
To remedy the mismatch between assumptions and realistic conditions, in this work we propose objectives which can serve as alternatives to \cref{eq:metaobj} of vanilla MAML. These are rooted in an interpretation of MAML within a decision-theoretic perspective \citep[ch.\ 1.3]{bickel2015mathematical}, which we outline in what follows. The quantity of interest we aim at learning is the neural parameters $\vtheta$. %
Therefore, the action space for a classification task assigning labels $y \in \mathcal{Y}$ to inputs $\xx \in \mathcal{X}$ is $\mathcal{A} = \{f_\vtheta : \mathcal{X} \rightarrow \mathcal{Y} \}$. The risk function is in turn a function $\mathcal{R} : \mathcal{F} \times \mathcal{A} \rightarrow \mathbb{R}^{+}$, which is the loss incurred by taking an action in $\mathcal{A}$ (making a prediction with a specific configuration of neural parameters) when the `state of nature', the true function, is $f \in \mathcal{F}$. In the case of MAML, this is represented by the language-specific inner loop loss $\mathcal{L}_{\mathcal{T}_i}(\cdot)$ in \cref{eq:metaobj}.

The decision for the optimal action given the sample space, the function $\delta : \mathcal{X} \times \mathcal{Y} \rightarrow \mathcal{A}$, is usually determined via gradient descent optimisation for a neural network.
The optimal action, however, may vary depending on the language, which results in multiple possible `states of nature'. Usually, there is no procedure $\delta$ whose loss is inferior to all others, such that:
\begin{equation}
   \nexists\delta \; \mathcal{L}(\mathcal{T}_i, \delta) < \mathcal{L}(\mathcal{T}_i, \delta^\prime) \; \forall \mathcal{T}_i \in \mathcal{T}, \delta \neq \delta^\prime
\end{equation}
\noindent
Therefore, decision functions have to be compared based on a global criterion rather than in a pair-wise fashion between languages.
As previously anticipated, \cref{eq:metaobj} minimizes the \textit{expected} risk across languages, for an arbitrary choice of prior $p(\mathcal{T})$. In decision theory, a decision $\delta^\star$ with this property is called \textit{Bayes criterion}.%

\subsection{Alternative Criteria}
\label{ssec:mmnpmaml}
There exist alternative criteria to the Bayes criterion that are more justified in a setting that entails transfer between non-i.i.d.\ domains. Rather than minimising the Bayes risk, in this work, we propose to adjust MAML to either minimise the maximum risk (minimax criterion) or to enforce constraints on the risk for a subset of languages (Neyman--Pearson criterion). This is likely to yield more robust predictions for languages that are outliers to the training distribution. As demonstrated in \cref{sec:skewlangdist}, this definition encompasses most of the world's languages.

\subsubsection{Minimax Criterion} Rather than the expected risk, the criterion could depend instead on the worst case scenario, i.e.\ the language for which the risk is \textit{maximum}. This requires to select such a language with $\textrm{max}$. As an alternative to reinforcement learning \citep{zhang2020worst}, to keep our model fully differentiable, we relax the operator by treating the choice of language as a categorical distribution $\mathcal{T}_i \sim \text{Cat}(\cdot \mid \lv\tau)$. The parameters $\lv\tau \in [0, 1]^{|\mathcal{T}|}, \sum_i \tau_i = 1$ consist of language probabilities and are learned in an adversarial fashion: 
\begin{equation} \label{eq:minimax}
    \min_{\vtheta} \max_{\mathcal{T}_i \sim \text{Cat}(\cdot \mid \lv\tau)} \mathcal{L}_{\mathcal{T}_i} (f_{\vtheta - \eta \nabla_\vtheta \mathcal{L}_{\mathcal{T}_i}(f_\vtheta, \mathcal{D}_{train})}, \mathcal{D}_{val})
\end{equation}
\noindent
\Cref{eq:minimax} can be interpreted as a two-player game between us (the scientists) and nature. We pick an action $\vtheta$. Then nature picks a language $\mathcal{T}_i \in p(\mathcal{T})$ for which the risk is maximum given our chosen action. Therefore, our goal becomes to minimise such risk.

\subsubsection{Neyman--Pearson Criterion}
As an alternative, we might consider minimising the expected risk, but subject to a guarantee that the risk does not exceed a certain threshold for a subset of languages. In practice, we may want to enforce a set of inequality constrains, so that we minimise \cref{eq:metaobj} subject to $\{\mathcal{L}_{\mathcal{T}_i} \leq r \; \forall \mathcal{T}_i \in \mathcal{C}\}$, where $r \in \mathbb{R}_{+}$ is a hyper-parameter. In general, $\mathcal{C} \subseteq \mathcal{T}$ can be any subset of the training languages; in practice, here we take $\mathcal{C} = \mathcal{T}$. Constrained optimisation is usually implemented through Lagrange multipliers, where we add as many new terms to the objective as we have constraints \citep[ch. 7]{bishop2006pattern}:
\begin{align} \label{eq:neymanpearson} \nonumber
    \min_{\vtheta} \max_{\lv{\lambda}} &\sum_{\mathcal{T}_i } \frac{1}{|\mathcal{T}|} \mathcal{L}_{\mathcal{T}_i}
    + \sum_{\mathcal{T}_i} \lambda_i (\mathcal{L}_{\mathcal{T}_i} - r) \\
    =  \min_{\vtheta} \max_{\lv{\lambda}} &\sum_{\mathcal{T}_i} \left(\frac{1}{|\mathcal{T}|} + \lambda_i\right) \mathcal{L}_{\mathcal{T}_i} - \lambda_i r
\end{align}
\noindent
where $\lv{\lambda}$ is a vector of non-negative Lagrange multipliers $\{\lambda_i \geq 0 \quad \forall \lambda_i \in \lv{\lambda}\}$ to be learned together with the parameters $\vtheta$, but adversarially.

Intuitively, if the risk for the estimated parameters $\vtheta$ lies in the permissible range, the constraints should become inactive $\{\lambda_i = 0 \quad \forall \lambda_i \in \lv\lambda\}$, i.e.\ each Lagrange multiplier should go towards 0. Otherwise, the solution should be affected by the constraints, which should keep $\vtheta$ from trespassing the boundary $\{\mathcal{L}(\vtheta)_{\mathcal{T}_i} = r \quad \forall \mathcal{T}_i \in \mathcal{T}\}$. In gradient-based optimisation, this unfolds as follows: the gradient of each $\lambda_i$ depends uniquely on $(\mathcal{L}_{\mathcal{T}_i} - r)$. Due to being maximised, the value of each $\lambda_i$ increases when the corresponding risk is above the threshold, and shrinks otherwise.
Incidentally, note that the Lagrangian multipliers at the critical point $\vtheta^\star$ are equal to the negative rate of change of $r$, as $\frac{\partial \mathcal{R}(\vtheta^\star)}{\partial r} = -\lv\lambda$. In other words, upon convergence $\lambda_i$ expresses how much we can decrease the risk in $\mathcal{T}_i$ as we increase the threshold.

\paragraph{Constrained Parameters}
The additional variables $\lv\tau$ and $\lv\lambda$, contrary to the neural parameters, are constrained in the values they can take. %
In neural networks, there are two widespread approaches to coerce variables within a certain range, viz.\ reparametrisation and gradient projection \citep{beck2003mirror}.\footnote{\url{https://vene.ro/blog/mirror-descent.html}} For simplicity's sake, we opt for the former, which just requires us to learn unconstrained variables and scale them with the appropriate functions. Thus, we redefine the above-mentioned variables as $\lv\tau \triangleq \textrm{softmax}(\lv\tau_u)$ and $\lv\lambda \triangleq \textrm{softplus}(\lv\lambda_u)$.

\section{Optimisation in 2-Player Games}
\label{sec:compopt}

Based on the formulation of Minimax MAML and Neyman--Pearson MAML in \cref{ssec:mmnpmaml}, both are evidently instances of two-player games. On one hand, the first agent minimises the risk with respect to $\vtheta$; on the other, the second agent maximises the risk with respect to $\lv{\tau}$ (for minimax) or  $\lv{\lambda}$ (for Neyman--Pearson). In other words, both optimise the same (empirical risk) function in \cref{eq:minimax} or \cref{eq:neymanpearson}, respectively, but with opposite signs. However, the first term of \cref{eq:neymanpearson} does not depend on $\lv{\lambda}$. Therefore, Minimax MAML is a zero-sum game, but not Neyman--Pearson MAML.

If the risk function were convex, the solution would be well-defined as the Nash equilibrium. But this is not the case for a non-linear function such as a deep neural network. Therefore, we resort to an approximate solution through optimisation. The simplest approach in this scenario is Gradient Descent Ascent (GDA), where the set of parameters of both players are optimised simultaneously through gradient descent for the first player and gradient ascent for the second player. With a slight abuse of notation, let us define %
$\mathcal{R} \triangleq \mathcal{R}(\vtheta_t, \lv\alpha_t)$, where $\lv\alpha_t$ stands for the adversarial parameters ($\lv\tau_t$ for Minimax and $\lv\lambda_t$ for Neyman--Pearson) at time $t$. Then the update rule is:
\begin{align} \label{eq:simgd1}
    \vtheta_{t+1} &= \vtheta_t - \eta \nabla_\vtheta \mathcal{R} \\
    \lv\alpha_{t+1} &= \lv\alpha_t + \eta \nabla_\lv\alpha \mathcal{R}  \label{eq:simgd2}
\end{align}
\noindent
for a learning rate $\eta \in \mathbb{R}$. \Cref{eq:simgd1,eq:simgd2} are equivalent to allowing each player to ignore the other's move and act as if it will remain stationary. This na\"ive assumption often leads to divergence or sub-par solutions during optimisation \citep{schafer2019competitive}.

\subsection{Symplectic Gradient Adjustment}
To overcome the limitations of GDA, several independent works \citep{balduzzi2018mechanics,letcher2019differentiable,gemp2018global} proposed to correct \cref{eq:simgd1,eq:simgd2} with an additional term. This consists of a matrix-vector product between the mixed second-order derivatives %
($\D_{\vtheta\lv\alpha}^2 \mathcal{R}$ and $\D_{\lv\alpha\vtheta}^2 \mathcal{R}$, respectively)\footnote{Here $\D^2_{\lv{w}\lv{z}} \mathcal{R}$ stands for the sub-matrix of the Hessian containing the derivative of the risk taken first with respect to $\lv{w}$ and then with respect to $\lv{z}$.} and the gradient of the risk with respect to the adversarial parameters ($\nabla_\lv\alpha \mathcal{R}$ and $\nabla_\vtheta \mathcal{R}$, respectively). The resulting optimisation algorithm, Symplectic Gradient Adjustment (SGA), updates parameters as follows:
\begin{align} \label{eq:lcgd1}
    \vtheta_{t+1} &= \vtheta_t - \eta \nabla_\vtheta \mathcal{R} - \eta^2 \D_{\vtheta\lv\alpha}^2 \mathcal{R} \; \nabla_\lv\alpha \mathcal{R}   \\
    \lv\alpha_{t+1} &= \lv\alpha_t + \eta \nabla_\lv\alpha \mathcal{R} - \eta^2 \D_{\lv\alpha\vtheta}^2 \mathcal{R} \; \nabla_\vtheta \mathcal{R}  \label{eq:lcgd2}
\end{align}
\noindent
Intuitively, the mixed second-order derivative represents the interaction between the players, and the adversarial gradient represents the opponent's move if they follow the simple GDA strategy. \citet{schafer2019competitive} cogently demonstrate how \cref{eq:lcgd1,eq:lcgd2} correspond to an approximation of the Nash equilibrium\footnote{A Nash equilibrium is a pair of strategies whose unilateral modification cannot result in loss reductions.} of a local bi-linear approximation (with quadratic regulariser) of the underlying game dynamics.

In practice, estimating the above-mentioned products is tedious because of their space and time complexity. Therefore, we resort to an approximation known as \textit{Hessian-vector product} \citep{pearlmutter1994fast}. For the third term of \cref{eq:lcgd1}:
\begin{align} \label{eq:hesvecprod}
    &\D^2_{\vtheta\lv\alpha} \mathcal{R}(\vtheta, \lv\alpha) \nabla_\lv\alpha \mathcal{R}(\vtheta, \lv\alpha) \nonumber \\
    = &\frac{\partial}{\partial h} \nabla_\vtheta \mathcal{R}(\vtheta, \lv\alpha + h \nabla_\lv\alpha \mathcal{R}(\vtheta, \lv\alpha)) \biggr\rvert_{h = 0}
\end{align}
\noindent
And similarly for the matrix product term in \cref{eq:lcgd2}, by swapping $\vtheta$ and $\lv\alpha$ in \cref{eq:hesvecprod}.

\subsection{Adaptive Learning Rate and Momentum}
While SGA may provide a more appropriate optimisation framework for competitive games, it still lacks several defining features of optimisers that accelerate convergence, such as first-order momentum and adaptive learning rate (second-order momentum). Therefore, we modify the update rule in \cref{eq:lcgd1,eq:lcgd2} to include both of these. Our starting point is Adam \citep{DBLP:journals/corr/KingmaB14}. The changes we apply are the following (also illustrated in \cref{alg:alcgd}):

\definecolor{vred}{HTML}{AE1C28}
\definecolor{vblue}{HTML}{21468B}

\renewcommand\algorithmicdo{}
\newcommand{\mathcolorbox}[2]{\colorbox{#1}{$\displaystyle #2$}}

\begin{algorithm}[t]
\caption{Adaptive Symplectic Gradient Adjustment (ASGA)}
   \label{alg:alcgd}
   \begin{algorithmic}[1]
\REQUIRE{$\eta \in \mathbb{R}_+$}: Learning rate
\REQUIRE{$\beta_1, \beta_2 \in [0, 1)$}: Decay rates
\REQUIRE{$\vtheta_0, \lv\alpha_0$}: Initial parameter values
\REQUIRE{$\mathcal{R} \triangleq \mathcal{R}(\vtheta_{t-1}, \lv\alpha_{t-1}): \mathbb{R}^{|\vtheta| + |\lv\alpha|} \rightarrow \mathbb{R}$}
\STATE{$\lv{m}_{0} \gets \lv{0}$ Initialise first moments}
\STATE{$\lv{v}_{0} \gets \lv{0}$ Initialise second moments}
\STATE{$t \gets 0$ Initialise time step}
\WHILE{$\vtheta_t, \lv\alpha_t$ not converged}{
\STATE $t \gets t + 1$
\STATE $\lv{g}_{\vtheta, t} \gets \nabla_\vtheta \mathcal{R} \textcolor{vred}{+ \eta D_{\vtheta\lv\alpha} \mathcal{R} \; \nabla_\lv\alpha \mathcal{R}}$
\STATE $\textcolor{vblue}{\lv{g}_{\lv\alpha, t} \gets \nabla_\lv\alpha \mathcal{R}} \textcolor{vred}{- \eta D_{\lv\alpha\vtheta} \mathcal{R} \; \nabla_\vtheta \mathcal{R}}$
\STATE $\textcolor{vblue}{\lv{g}_{t} \gets \lv{g}_{\vtheta, t} \oplus \lv{g}_{\lv\alpha, t}}$
\STATE $\lv{m}_{t} \gets {\beta_1 \, \lv{m}_{t-1} + (1 - \beta_1) \, \lv{g}_{t}}$
\STATE $\lv{v}_{t} \gets \beta_2 \, \lv{v}_{t-1} + (1 - \beta_2) \, \lv{g}_{t}^2$
\STATE $\lv{{\hat{m}}}_{t} \gets \lv{m}_{t} \, / \, (1 - \beta_1^t)$
\STATE $\lv{{\hat{v}}}_{t} \gets {\lv{v}_{t} \, / \, (1 - \beta_2^t)}$
\STATE $\vtheta_{t} \gets \vtheta_{t - 1} - \eta \cdot \lv{{\hat{m}}}_{\vtheta, t} \, / \, (\sqrt{\lv{{\hat{v}}}_{\vtheta, t}} + \epsilon)$
\STATE $\textcolor{vblue}{\lv\alpha_{t} \gets \lv\alpha_{t - 1} + \eta \cdot \lv{{\hat{m}}}_{\lv\alpha, t} \, / \, (\sqrt{\lv{{\hat{v}}}_{\lv\alpha, t}} + \epsilon)}$
}
\ENDWHILE
\RETURN{$\vtheta_t, \lv\alpha_t$}
\end{algorithmic}
\end{algorithm}

\newcommand*\circled[2]{\tikz[baseline=(char.base)]{%
\node[circle,fill=#1,draw,opacity=0.66,scale=0.66] (char) {#2};}}

\def\cnt{\stepcounter{enumi}\arabic{enumi}}

\renewcommand{\labelenumi}{(\arabic{enumi})}
\begin{enumerate}
\itemsep0em 
\item[\circled{vred}{\cnt}] The current difference (lines 6--7) is adjusted with the terms introduced in \cref{eq:lcgd1,eq:lcgd2} by \citet{schafer2019competitive}.
\item[\circled{vblue}{\cnt}] The exponentially decayed, unbiased estimates of the expectations over mean and standard deviation are computed similarly to Adam. However, note that, in line 14, the update of the adversarial parameters corresponds to an ascent (rather than a descent).
\end{enumerate}

\noindent
This results in a novel optimiser, Adaptive Symplectic Gradient Adjustment (ASGA). We employ ASGA in our experiments to optimise the objectives of Minimax MAML and Neyman--Pearson MAML, as it enables a fair comparison with Adam-optimised Bayes MAML.

\section{Experiments}

We now outline the main experiments of our work on multilingual NLP. We evaluate our methods on part-of-speech (POS) tagging, a sequence labelling tasks, and question answering (QA), a natural language understanding task. 

We focus on POS given its ample coverage of languages and its frequent use as a benchmark for resource-poor NLP \citep{das2011unsupervised,ponti2020parameter}. In fact, cross-lingual transfer in sequence labelling tasks was demonstrated to be the most challenging, as knowledge of linguistic structure is more language-dependent than semantics \citep{hu2020xtreme}. However, we also include QA to illustrate the generality of our methods for cross-lingual NLP. In this task, given the gold passage and a question, the system has to predict the beginning and end positions of a single contiguous span containing the answer. 

\vspace{1.3mm}
\noindent \textbf{Data.}
POS data are sourced from the Universal Dependencies (UD) treebanks\footnote{\url{https://universaldependencies.org/}} \cite{11234/1-3226} %
and QA data from the `gold passage' variant of TyDiQA \citep{tydiqa}.\footnote{\url{https://github.com/google-research-datasets/tydiqa}} We retain the original training, development, and evaluation sets of UD. %
In TyDiQA, we use the original development set for evaluation.\footnote{This is necessary as we need to access this set to simulate few-shot learning, but the original evaluation set is not public.} For meta-learning, $\mathcal{D}_{train}$ and $\mathcal{D}_{val}$ examples are both obtained from disjoint parts of the training set.

We aim to create a partition of languages between training and evaluation that corresponds to the most realistic scenario in deploying NLP technology on resource-poor languages spoken around the world. Therefore, we reserve for evaluation all language isolates and languages with at most 2 family members in each dataset. We use all the remaining languages in the dataset for training. Therefore, for POS, the evaluation set spans 16 treebanks (14 languages, 11 families) and the training set 99 treebanks; 
QA comprises 9 languages (7 families). We hold out 4 of them in turn for evaluation (except English) and use the rest for training. We provide the full list of languages in \cref{app:langpart}.

\vspace{1.3mm}
\noindent \textbf{Training.} 
In all tasks, we train a neural network consisting of two stacked modules: an encoder and a classifier. The encoder is a 12-layer, 768-hidden unit, 12-head Transformer initialised with multilingual BERT Base (mBERT), which was pre-trained on cased text from 104 languages.\footnote{\url{https://github.com/google-research/bert/blob/master/multilingual.md}} The classifier is a single affine layer for TyDiQA and a 2-layer Perceptron (with 1024 hidden units) for POS tagging. The combined parameters of the encoder and classifier correspond to $\vtheta$ from \cref{sec:robustmaml}. 

These are meta-learned via Meta-SGD \citep{li2017meta}, a first-order MAML variant where each parameter is assigned a separate inner-loop learning rate $\eta$. Moreover, each $\eta$ is trained end-to-end based on the outer-loop loss (such as \cref{eq:metaobj} for the Bayes criterion).\footnote{We implement Meta-SGD with the \texttt{learn2learn} package \citep{Arnold2020-ss}.} Similar to \citet{bansal2019learning}, to avoid an explosion in the number of parameters, we assign a per-layer learning rate (rather than per-parameter). To avoid overfitting, we employ both dropout (with a probability of $0.2$) and early stopping (with a patience of $10$).
For the Neyman-Pearson formulation, we set $r = 0.1$ as a threshold for all language-specific losses.\footnote{We also experimented with a dynamic threshold which corresponded to the average language-specific loss of the last 10 episodes. However, this yielded sub-par results.} The parameters $\lv\tau$ and $\lv\lambda$ were initialized uniformly as $\frac{1}{|\mathcal{T}|}$. Complete details of the hyper-parameters for all settings are given in Appendix~\ref{app:hyper}.

\vspace{1.3mm}
\noindent \textbf{Methods.}
To assess the effectiveness of the proposed criteria and optimisers, we compare them with two competitive baselines, while maintaining the same underlying neural architecture:
(i) \textbf{J}: a joint multi-source transfer method where a model is trained on the concatenation of the datasets for all languages; (ii) \textbf{B:} the original MAML \citep{finn2017model} with Bayes criterion and uniform prior. Our choice of baselines is justified by the fact that these methods (or variations thereof) are currently state of the art for the tasks of POS and QA, as well as other innumerable NLP applications \citep{wu2019beto,nooralahzadeh-etal-2020-zero,ponti2020parameter}. In addition, we evaluate the following combinations: (iii) \textbf{MM:} MAML with a minimax criterion, optimised with GDA; 
(iv) \textbf{NP:} MAML with a Neyman--Pearson (constrained) criterion, optimised with GDA;
(v) \textbf{MM+:} MAML with a minimax criterion, optimised with ASGA; and
(vi) \textbf{NP+:} MAML with a Neyman--Pearson criterion, optimised with ASGA.

\vspace{1.3mm}
\noindent \textbf{Evaluation.} For each evaluation language in a given task, we randomly sample $k \in \lbrace 0, 5, 10, 20 \rbrace$ examples from the evaluation data as the support set (for adaptation) and the rest of the examples as the query set (for testing). When $k>0$, we repeat the evaluation 100 times and report the following average metrics: (i) F$_1$ score for POS tagging, and (ii) exact-match (EM) and F$_1$ scores for QA.\footnote{We refer the reader to \citet{rajpurkar-etal-2016-squad} for a precise definition of these metrics.} %
Due to lack of space, we only report the average mean and standard deviation across languages for each model described above. %

\section{Results and Discussion}

\definecolor{Gray}{gray}{0.92}

\begin{table}[t]
\small
\centering
\begin{tabular}{l|cccc}
\toprule
k $\triangleright$ & 0 & 5 & 10 & 20 \\
\midrule
\rowcolor{Gray}
\multicolumn{5}{c}{F$_1$ Score} \\
J & 51.01 & 62.96$\pm$2.5 & 66.00$\pm$1.9 & 68.66$\pm$1.7 \\
B & 51.50 & 63.87$\pm$2.8 & 67.03$\pm$2.1 & 69.46$\pm$1.8 \\
\midrule
MM & 51.82 & 63.67$\pm$2.7 & 66.88$\pm$2.0 & 69.55$\pm$1.8 \\
NP & 51.68 & 63.84$\pm$2.9 & 67.13$\pm$2.1 & 69.65$\pm$1.9 \\
MM+ & 52.46 & \textbf{64.71$\pm$2.9} & \textbf{67.89$\pm$2.3} & \textbf{70.25$\pm$2.0} \\
NP+ & \textbf{53.05} & 64.26$\pm$2.6 & 67.57$\pm$2.1 & 69.98$\pm$1.9 \\
\bottomrule
\end{tabular}
\caption{F$_1$ scores for POS tagging in UD across different \textit{k}-shots. We report the mean and standard deviation across 16 treebanks.}
    \label{tab:unweighted-pos-results}
\end{table}

\begin{table}[t]
\small
\centering
\begin{tabular}{l|cccc}
\toprule
k $\triangleright$ & 0 & 5 & 10 & 20 \\
\midrule
\rowcolor{Gray}
\multicolumn{5}{c}{Exact Match} \\
J & 46.76 & 49.53$\pm$3.7 & 51.54$\pm$2.9 & \textbf{53.51$\pm$2.4} \\
B & 46.60 & 48.41$\pm$3.4 & 50.24$\pm$2.9 & 52.02$\pm$2.6 \\
\midrule
MM & \textbf{48.33} & \textbf{50.08$\pm$3.4} & \textbf{51.68$\pm$2.9} & \textbf{53.49$\pm$2.4} \\ 
NP &  46.71 & 49.24$\pm$3.3 & 50.95$\pm$2.9 & 52.76$\pm$2.4\\
MM+ & 46.87 & 47.74$\pm$3.8 & 49.42$\pm$3.4 & 51.40$\pm$2.5 \\
NP+ & 48.02 & 48.77$\pm$3.9 & 50.75$\pm$3.1 & 52.66$\pm$2.6 \\
\midrule
\rowcolor{Gray}
\multicolumn{5}{c}{F$_1$ Score} \\
J & 61.66 & 63.75$\pm$3.3 & 65.39$\pm$2.3 & 67.01$\pm$1.9 \\
B & 62.51 & 63.29$\pm$3.2 & 64.87$\pm$2.5 & 66.31$\pm$2.1 \\
\midrule
MM & \textbf{63.06} & \textbf{64.37$\pm$3.1} & \textbf{65.83$\pm$2.6} & \textbf{67.45$\pm$2.1} \\
NP & 61.89 & 63.84$\pm$2.9 & 65.23$\pm$2.6 & 66.88$\pm$1.9 \\
MM+ & 62.10 & 62.63$\pm$3.2 & 64.11$\pm$2.9 & 65.89$\pm$2.1 \\
NP+ & 62.75 & 62.98$\pm$3.6 & 64.77$\pm$2.9 & 66.57$\pm$2.2 \\
\bottomrule
\end{tabular}
\caption{Results for QA in TyDiQA across different \textit{k}-shots. We report the mean and standard deviation across 8 languages of the exact match score (above) and the F$_1$ score (below).}
    \label{tab:unweighted-tydiqa-results-exact}
    \vspace{-0.4cm}
\end{table}

We report the results for POS tagging in \cref{tab:unweighted-pos-results} and for QA in \cref{tab:unweighted-tydiqa-results-exact}. These include mean and standard deviation across languages. Note that, in this case, the standard deviation is by no means an interval for statistical significance, but rather reflects the heterogeneity among the evaluation languages. In what follows, we address a series of questions in the light of these figures.

\vspace{1.3mm}
\noindent 
\textbf{Baselines.} MAML and joint multi-source transfer are both strong contenders as state-of-the-art methods for cross-lingual transfer, but which one is better? By comparing J and B rows, no definite response emerges in our experiments. While MAML outperforms its competitor in POS tagging, it lags behind in QA. We speculate that the larger pool of training languages available in POS tagging (22 times more than QA) endows meta-learning with better generalisation capabilities. Both methods, however, surpass single-source transfer from English SQuAD \citep{rajpurkar-etal-2016-squad} in the zero-shot setting by a large margin: \citet{tydiqa} report 56.4 F$_1$ score in average for TyDiQA, which is 6.66 points below our best model.

\vspace{1.3mm}
\noindent 
\textbf{Criteria.}
The minimax and Neyman-Pearson criteria both improve over the Bayes criterion baseline, although the latter more sporadically. Compared to the B rows, MM+ achieves gains for every \textit{k} in POS tagging, with $0.94$ points of margin at $k=0$ and $0.79$ at $k=20$. The same holds for MM in QA, with margins that span from $1.73$ at $k=0$ to $1.47$ at $k=20$ in the Exact Match metric, and from $0.55$ at $k=0$ to $1.14$ at $k=20$ in F$_1$ score. Therefore, Minimax MAML is remarkably consistent in outperforming the baselines, although the gains are sometimes significant, sometimes only marginal. This is also reflected in language-specific performances, available in \cref{tab:res_ud2} and \cref{tab:res_qa2} in the Appendix. For POS tagging, the F$_1$ scores of only 2 languages (Indonesian and Naija) moderately decrease, whereas the rest of the 14 languages show improvements. 

Incidentally, it may be worth noting that we did not perform any large-scale search over hyper-parameters like $\lv\tau$ and $\lv\lambda$ initialisations, the threshold $r$, or differential learning rates for maximised and minimised parameters. Therefore, these early results are amenable to improve even further in the future. This lends credence to our proposition that minimax and Neyman--Pearson criteria are more suited for out-of-distribution transfer to outlier languages.

\begin{figure*}
    \centering
    \includegraphics[width=\textwidth]{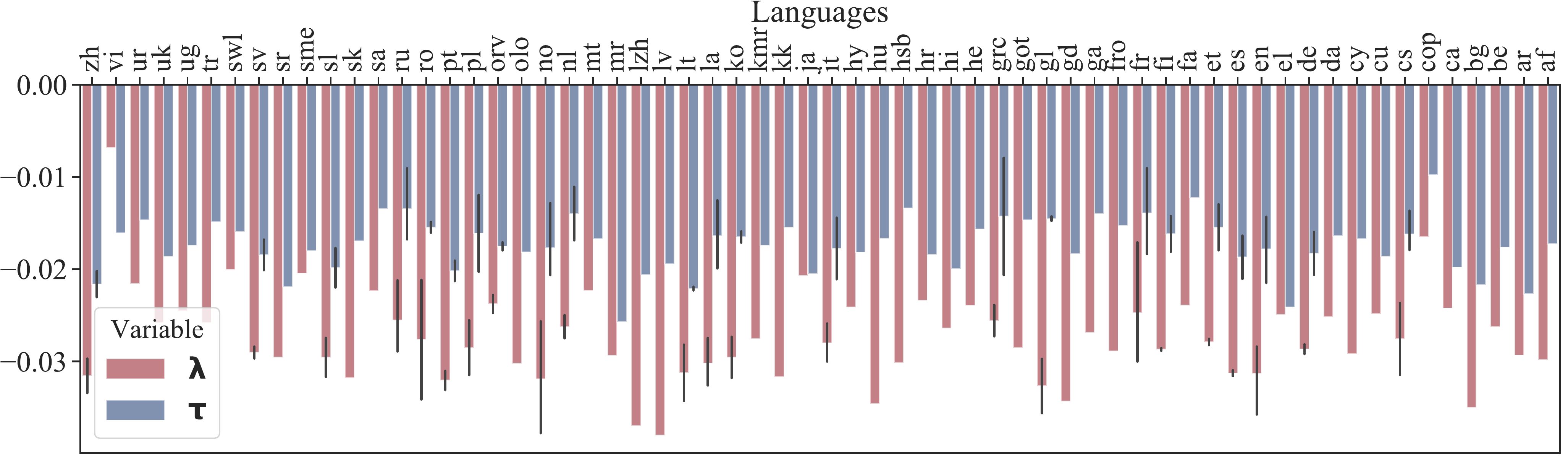}
    \caption{Unconstrained values of $\lv\tau_u$ and $\lv\lambda_u$ upon convergence in MM+ and NP+ models for POS tagging.}
    \label{fig:tau-lambda-pos}
    \vspace{0.3cm}
\end{figure*}

\begin{table}[t]
\small
\centering
\begin{tabular}{l|cccc}
\toprule
k $\triangleright$ & 0 & 5 & 10 & 20 \\
\midrule
\rowcolor{Gray}
\multicolumn{5}{c}{F$_1$ Score} \\
J & 14.34 & 33.32 & 37.52 & 40.83 \\
B & 24.11 & 35.03 & 40.38 & 44.92 \\
\midrule
MM & 20.41 & 37.61 & 43.00 & 45.83 \\
NP & \textbf{26.81} & 39.23 & 42.70 & 45.25 \\
MM+ & 16.42 & 37.41 & 43.57 & 45.21 \\
NP+ & 22.55 & \textbf{39.95} & \textbf{45.41} & \textbf{48.12} \\
\bottomrule
\end{tabular}
\caption{The minimum F$_1$ scores of our models across languages, for POS tagging.}
    \label{tab:worst-f1}
\end{table}

\begin{table}[t]
\small
\centering
\begin{tabular}{l|cccc}
\toprule
k $\triangleright$ & 0 & 5 & 10 & 20 \\
\midrule
\rowcolor{Gray}
\multicolumn{5}{c}{Exact Match} \\
J & 42.33 & \textbf{45.97} & 47.22 & \textbf{49.47} \\
B & \textbf{42.75} & 44.58 & 46.44 & 48.24 \\
\midrule
MM & 41.01 & 45.33 & \textbf{47.59} & 49.21 \\
NP & 40.39 & 44.89 & 46.40 & 48.80 \\
MM+ & 41.01 & 41.87 & 45.32 & 47.11 \\
NP+ & 37.44 & 42.92 & 46.30 & 48.88 \\
\rowcolor{Gray}
\multicolumn{5}{c}{F$_1$ Score} \\
J & 52.43 & 59.27 & 59.88 & 62.11 \\
B & 51.10 & \textbf{59.60} & 60.64 & 62.10 \\
\midrule
MM & 53.10 & 59.21 & \textbf{61.86} & 63.43 \\
NP & 52.83 & 59.31 & 60.03 & 61.84 \\
MM+ & 51.93 & 57.91 & 59.86 & 61.52 \\
NP+ & \textbf{53.96} & 57.21 & 61.74 & \textbf{63.55} \\
\bottomrule
\end{tabular}
\caption{The minimum Exact Match and F$_1$ scores of our models across languages, for QA.}
    \label{tab:worst-f1-qa}
    \vspace{-0.2cm}
\end{table}

\vspace{1.3mm}
\noindent \textbf{Optimiser.}
The results for the proposed optimiser ASGA (\cref{alg:alcgd}) are favourable in comparison to Gradient Descent Ascent via Adam \citep{DBLP:journals/corr/KingmaB14} for POS tagging; on the other hand, the opposite trend is observed for QA. Therefore, future investigations are required to shed further light on modifications such as the Symplectic Gradient Adjustment. A tentative explanation of such discrepancy could be the disproportionate number of training languages available in either task.

To get insights into the game dynamics of the adversarial criteria, we plot the unconstrained values for $\lv\tau_u$ and $\lv\lambda_u$ upon convergence in \cref{fig:tau-lambda-pos}. Interestingly, both variables appear to follow the same profile of peaks and troughs; therefore, as expected, languages chosen adversarially in MM have also higher Laplace multipliers in NP. To this group belong for instance languages with rare scripts (e.g. Coptic) or with no relatives in the training languages (e.g.\ Vietnamese). As a final note, we remark that the proposed criteria and optimiser are in principle more general than NLP and could facilitate transfer in other fields. While this thread of research transcends the scope of our work, we illustrate an example for regression in \cref{app:extraexp}.

\vspace{1.3mm}
\noindent 
\textbf{Minimum Scores across Languages.}
In addition to the \textit{average} cross-lingual performance, we also report the \textit{minimum} cross-lingual performance for POS tagging in \cref{tab:worst-f1} and for QA in \cref{tab:worst-f1-qa}. This corresponds to the lowest score achieved across all evaluation languages. For POS tagging, we observe that NP and NP+ outperform J and B by 7-12 and 2-5 F1 points, respectively. This reveals that worst-case and constrained risk minimisation drastically uplifts the scores for the most disadvantaged language. Nevertheless, the opposite trend is observed for QA: MM(+) and NP(+) do not alter the minimum score with respect to the F$_1$ metric, and even degrade it with respect to the exact-match metric. Again, we conjecture that these mixed findings may depend on the different amount and distribution of the training languages in the corresponding datasets: UD offers greater language coverage than TyDiQA, which gives better guidance.

\section{Related Work}

\ignore{
In the space of NLP, meta-learning, particularly MAML, is employed to use high-resource tasks or languages to find an optimal initialisation which makes learning efficient for a low-resource task or language using only a small number of examples.
For example, \citet{gu-etal-2018-meta} meta-train on translating high-resource languages to English and then adapt to low-resource to English translation.
Related ideas have been explored for cross-lingual inference, QA and entity recognition \cite{nooralahzadeh-etal-2020-zero,wu2020enhanced}.
We differ from these works in two aspects: 1) at meta-train time, we do not limit ourselves to high-resource languages given that many world languages are low-resource languages;
2) these methods use standard MAML minimising the expected risk, whereas we adapt to high-risk languages, a realistic setting since at test-time many world languages are likely to be outliers (see Section 2).
Our setting is robust to distribution shifts at meta-train and meta-test.
}

MAML is a cutting-edge method for cross-lingual transfer in several NLP tasks \citep[\textit{inter alia}]{gu-etal-2018-meta,nooralahzadeh-etal-2020-zero,wu2020enhanced,li-etal-2020-learn}. However, in all these experiments, the model is adopted in its standard formulation, minimising the expected risk. Therefore, its performance is prone to suffer in outlier languages. Moreover, the assumptions underlying our proposed variants are different from other instances of robust optimisation in NLP \cite{globerson06nightmare,oren-etal-2019-distributionally}. In particular, the target language distributions are not explicitly treated as subspaces or covariate shifts of source languages. In separate fields such as vision, previous attempts at worst-case-aware meta-learning include \citet{collins2020task}, who use a Euclidean version of the robust stochastic mirror-prox algorithm, and \citet{wang-etal-2020-balancing}, who rely on reinforcement learning. Our formulation is both fully differentiable and broader, as the decision-theoretic interpretation admits alternative criteria for MAML. What is more, 
to our knowledge we are the first to successfully augment MAML with minimax criteria in cross-lingual NLP and with Neyman--Pearson criteria in general.

\section{Conclusions}
To perform cross-lingual transfer to low-resource languages, under a decision-theoretic interpretation Model-Agnostic Meta-Learning (MAML) minimises the expected risk across training languages. Generalisation then relies on the evaluation languages being identically distributed. However, this assumption is incongruous for cross-lingual transfer in realistic scenarios. Therefore, we propose more appropriate training objectives that are robust to out-of-distribution transfer: Minimax MAML, where worst-case risk is minimised by learning an adversarial distribution over languages; and Neyman--Pearson MAML, where constraints are imposed on language-specific losses, so that they remain below a certain threshold. From a game-theoretic perspective, both of these variants consist of 2-player competitive games. Therefore, we also explore adaptive optimisers that take into account the underlying game dynamics. The experimental results on zero-shot and few-shot learning for part-of-speech tagging and question answering, whose datasets span tens of typologically diverse languages, confirm that in several settings the proposed criteria are superior to both vanilla MAML and transfer from multiple source languages.

\section*{Acknowledgements}
We thank the reviewers for their valuable feedback. Rahul Aralikatte and Anders S{\o}gaard are funded by a Google Focused Research Award.

\bibliography{references}
\bibliographystyle{acl_natbib}

\clearpage
\appendix

\begin{figure}[t!]
  \centering
  
  \tikz{ %
  \node[obs] (x) {$\lv{x}_{ij}$};%
     \node[latent,left=of x] (r) {$\lv{\varphi}_i$}; %
     \node[latent,left=of r] (f) {$\lv{\vartheta}$}; %
     \plate [inner sep=.3cm] {plate1} {(x)} {$N$}; %
     \tikzset{plate caption/.append style={below=0pt and 0pt of #1.south west}}
     \plate [inner sep=.6cm] {plate2} {(x) (r)} {$\mathcal{T}$}; %

     \edge {r} {x}
     \edge {f} {r}
  }
  \caption{Bayesian graphical model of MAML, where the variable $\lv\varphi_i$ is parameterised as $\vtheta - \eta \nabla_\vtheta \mathcal{L}_{\mathcal{T}_i}(f_\vtheta, \mathcal{D}_{train})$.}
  \label{fig:hierbayes2}
\end{figure}

\begin{figure*}[b]
    \centering
    \includegraphics[width=.95\textwidth]{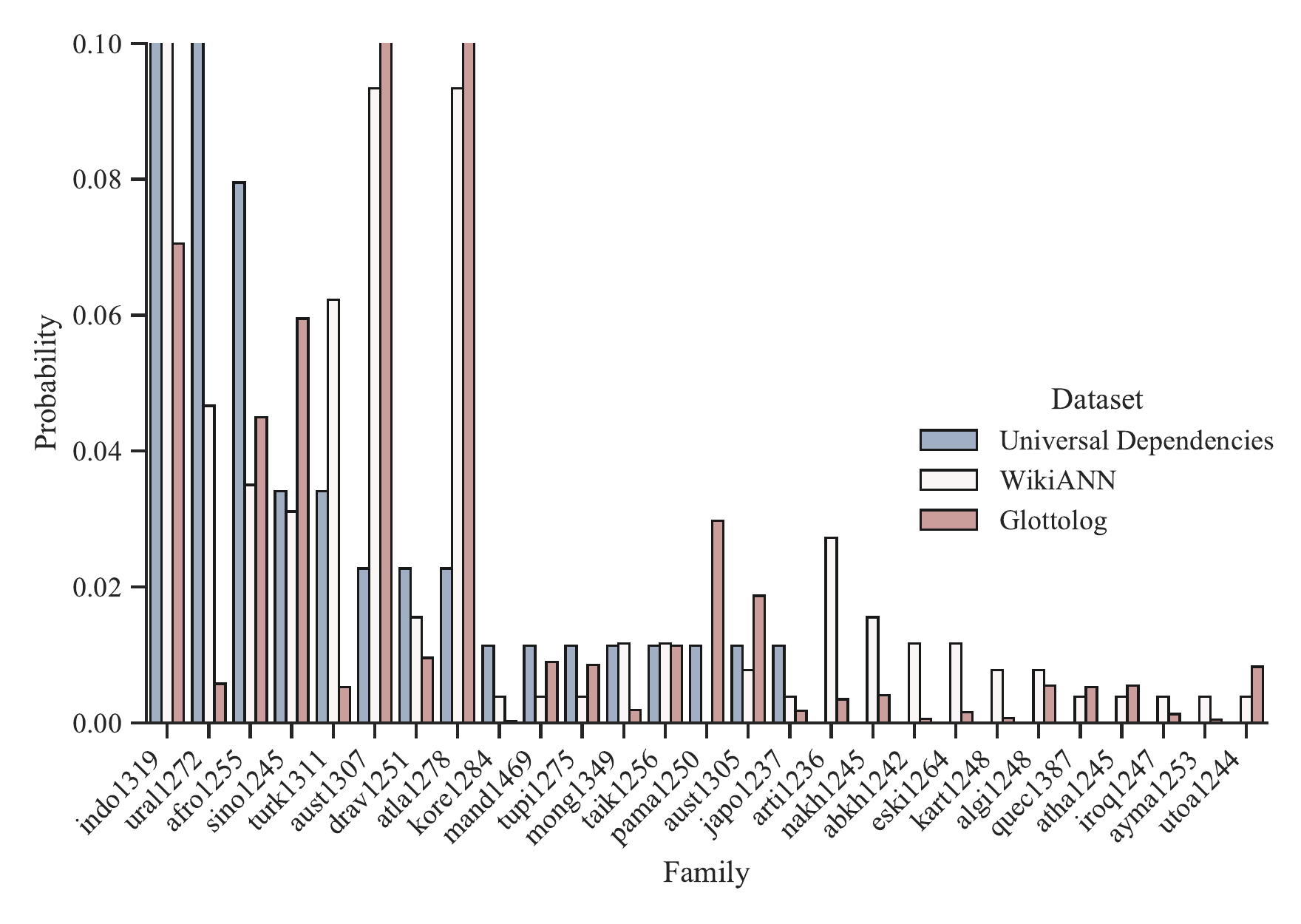}
    \caption{Empirical distribution of languages across families in 2 datasets (WikiANN and UD) and in the world, according to Glottolog. The families shown are a subset \{(WikiANN $\cup$ Universal Dependencies) $\cap$ Glottolog\}. The y-axis is truncated for the sake of clarity.}
    \label{fig:familydist}
\end{figure*}

\section{Language Partitions}
\label{app:langpart}
The languages from the following families in UD are held out for evaluation (16 treebanks, 14 languages in total): Northwest Caucasian (Abaza), Mande (Bambara), Mongolic (Buryat), Basque, Tupian (Mbya Guarani), Creole (Naija), Tai--Kadai (Thai), Pama--Nyungan (Warlpiri), Austronesian (Indonesian, Tagalog), Dravidian (Tamil, Telugu), Niger-Congo (Wolof, Yoruba). %
As all 8 languages in TiDiQA belong to families with at most 2 members in the dataset, we randomly create two partitions: in the former, Finnish, Korean, Bengali, and Arabic are used for evaluation, and the others for training; in the latter, Russian, Indonesian, Telugu, and Swahili are used for evaluation, and the others for training.

\section{Hyperparameter Setting}
\label{app:hyper}
\paragraph{POS Tagging.} 
For POS tagging: (i) the batch size was $32$, (ii) the maximum sequence length was $128$, (iii) the number of epochs was $20$, with a patience limit of $10$, (iv) both outer and inner learning rates were $5\times 10^{-5}$, (v) the number of episodes per iteration was $32$, (vi) the number of inner loops per outer update was $4$, (vii) the number of shots ($k$) during training was $30$, and (viii) the hidden layer dropout probability for the classifier was $0.2$.

\paragraph{QA.} (i) the batch size and $k$ were reduced to $12$ due to memory constraints, (ii) the maximum context length was $336$, and the document stride was $128$, (iii) the maximum question length was $64$, (iv) the inner and outer learning rates were $3\times 10^{-5}$.

For all J baselines, we used a uniform language sampler, since proportional sampling performed worse. As an optimiser, we chose Adam with a learning rate of $5\times 10^{-5}$, a weight decay of $0.1$; we clipped the gradient to a maximum norm of $5.0$. For all MAML models, we performed 4 updates in the inner loop, both during training and fast adaptation (few-shot learning). We ran our experiments on a 48GB NVIDIA Quadro RTX 8000 GPU with Turing micro-architecture. Each run took approximately 2 hours for training and 3 hours for few-shot learning and evaluation.

\begin{table*}[p]
\centering
\small
\begin{tabular}{r|c|cccccc}
\toprule 
Dataset & k & J & B & MM & NP & MM+ & NP+ \\
\midrule
\multirow{4}{*}{abq\_atb} & 0 & 14.34 & 24.11 & 20.41 & 26.81 & 16.42 & 22.55 \\
 & 5 & 33.32$\pm$5.58 & 35.03$\pm$5.85 & 37.61$\pm$4.57 & 39.23$\pm$5.41 & 37.41$\pm$7.3 & 39.95$\pm$5.93 \\
 & 10 & 37.52$\pm$2.28 & 40.38$\pm$5.75 & 43$\pm$3.63 & 42.7$\pm$5.3 & 43.57$\pm$7.31 & 46.56$\pm$4.91 \\
 & 20 & 40.83$\pm$6.53 & 44.92$\pm$7.08 & 45.83$\pm$5.59 & 45.25$\pm$7.26 & 45.21$\pm$9.4 & 48.12$\pm$7.19 \\
\multirow{4}{*}{bm\_crb} & 0 & 29.56 & 30.85 & 29.2 & 28.57 & 30.44 & 30.22 \\
 & 5 & 45.6$\pm$3.47 & 50.83$\pm$3.33 & 46.04$\pm$3.95 & 45.14$\pm$3.73 & 48.2$\pm$3.74 & 48.32$\pm$3.63 \\
 & 10 & 49.75$\pm$1.23 & 54.4$\pm$2.73 & 50.35$\pm$2.7 & 50.01$\pm$2.74 & 51.65$\pm$2.87 & 51.28$\pm$3.4 \\
 & 20 & 54.03$\pm$1.52 & 57.53$\pm$1.68 & 53.12$\pm$2.16 & 53.39$\pm$1.85 & 54.38$\pm$1.85 & 53.89$\pm$2.08 \\
\multirow{4}{*}{bxr\_bdt} & 0 & 48.85 & 51.71 & 50.41 & 50.49 & 54.21 & 51.94 \\
 & 5 & 51.29$\pm$1.67 & 51.81$\pm$2.18 & 51.57$\pm$2.21 & 51.62$\pm$2.17 & 53.09$\pm$2.08 & 51.83$\pm$2.31 \\
 & 10 & 53.64$\pm$0.96 & 54.95$\pm$1.68 & 54.18$\pm$1.53 & 54.25$\pm$1.63 & 55.47$\pm$1.8 & 55.17$\pm$1.43 \\
 & 20 & 56.18$\pm$1.13 & 57.23$\pm$1.17 & 56.48$\pm$1.49 & 56.97$\pm$1.2 & 58.19$\pm$1.38 & 57.29$\pm$1.12 \\
\multirow{4}{*}{eu\_bdt} & 0 & 70.2 & 71.76 & 73.22 & 72.57 & 73.54 & 73.29 \\
 & 5 & 74.7$\pm$1.39 & 75.74$\pm$1.69 & 75.42$\pm$1.59 & 75.77$\pm$1.94 & 76.58$\pm$1.64 & 76.52$\pm$1.64 \\
 & 10 & 76.51$\pm$2.38 & 78.1$\pm$1.25 & 77.52$\pm$1.01 & 78.08$\pm$1.21 & 78.73$\pm$1.36 & 78.19$\pm$1.38 \\
 & 20 & 78.52$\pm$0.67 & 80.09$\pm$0.87 & 79.47$\pm$0.84 & 80.01$\pm$0.76 & 80.69$\pm$0.91 & 80.24$\pm$0.78 \\
\multirow{4}{*}{gun\_thomas} & 0 & 32.06 & 35.72 & 33.91 & 31.97 & 33.87 & 33.84 \\
 & 5 & 40.65$\pm$2.27 & 42.62$\pm$3.05 & 43.28$\pm$2.64 & 42.32$\pm$2.45 & 43.12$\pm$2.63 & 42.46$\pm$2.37 \\
 & 10 & 44.06$\pm$0.99 & 45.65$\pm$2.33 & 45.92$\pm$2.59 & 45.23$\pm$2.31 & 46.98$\pm$2.49 & 45.41$\pm$2.25 \\
 & 20 & 46.46$\pm$2.07 & 47.96$\pm$2.11 & 50.34$\pm$2.3 & 48.15$\pm$2.09 & 50.67$\pm$2.15 & 48.44$\pm$1.74 \\
\multirow{4}{*}{id\_gsd} & 0 & 77.24 & 77.97 & 77.68 & 74.79 & 77.85 & 76.15 \\
 & 5 & 82.2$\pm$1.22 & 83.47$\pm$1.22 & 82.72$\pm$1.47 & 82.35$\pm$1.68 & 83$\pm$1.5 & 82.47$\pm$1.57 \\
 & 10 & 83.63$\pm$0.93 & 84.69$\pm$0.91 & 84.28$\pm$1.17 & 84.06$\pm$1.09 & 84.4$\pm$1.03 & 84.69$\pm$0.96 \\
 & 20 & 84.75$\pm$0.61 & 85.75$\pm$0.59 & 85.82$\pm$0.58 & 85.35$\pm$0.68 & 85.94$\pm$0.66 & 85.86$\pm$0.69 \\
\multirow{4}{*}{id\_pud} & 0 & 68.46 & 69.41 & 69.27 & 68.67 & 69.41 & 68.72 \\
 & 5 & 73.07$\pm$1.39 & 73.96$\pm$1.5 & 73.5$\pm$1.48 & 74.17$\pm$1.43 & 74.52$\pm$1.46 & 73.82$\pm$1.56 \\
 & 10 & 74.91$\pm$1.33 & 75.7$\pm$1.19 & 75.5$\pm$1.08 & 75.87$\pm$1.15 & 76.42$\pm$0.8 & 75.85$\pm$0.94 \\
 & 20 & 76.17$\pm$0.57 & 77.18$\pm$0.72 & 77.06$\pm$0.49 & 77.28$\pm$0.68 & 77.75$\pm$0.57 & 77.39$\pm$0.71 \\
\multirow{4}{*}{pcm\_nsc} & 0 & 61.97 & 40.78 & 45.77 & 40.76 & 41.21 & 56.83 \\
 & 5 & 78.17$\pm$1.58 & 77.87$\pm$1.27 & 77.42$\pm$1.67 & 76.48$\pm$1.74 & 77.33$\pm$1.55 & 77.71$\pm$1.78 \\
 & 10 & 80.06$\pm$1.24 & 79.28$\pm$1.25 & 78.96$\pm$1.1 & 78.41$\pm$1.1 & 78.71$\pm$0.94 & 80.03$\pm$1.37 \\
 & 20 & 81.61$\pm$0.85 & 80.6$\pm$0.81 & 80.17$\pm$0.8 & 79.97$\pm$1 & 80.13$\pm$0.72 & 81.99$\pm$0.99 \\
\multirow{4}{*}{ta\_ttb} & 0 & 55.65 & 56.31 & 58.12 & 58.47 & 60.18 & 55.93 \\
 & 5 & 72.29$\pm$2.03 & 72.39$\pm$2.21 & 71.37$\pm$1.7 & 72.28$\pm$2.46 & 72.34$\pm$2.13 & 70.19$\pm$2.3 \\
 & 10 & 74.73$\pm$2.27 & 75.36$\pm$1.47 & 73.7$\pm$1.36 & 75.51$\pm$1.54 & 75.11$\pm$1.47 & 73.69$\pm$1.73 \\
 & 20 & 76.23$\pm$1.19 & 77.56$\pm$1.38 & 75.75$\pm$1.39 & 77.83$\pm$1.33 & 77.44$\pm$1.3 & 76.29$\pm$1.49 \\
\multirow{4}{*}{te\_mtg} & 0 & 75.21 & 75.87 & 77.49 & 75.43 & 76.28 & 76.29 \\
 & 5 & 76.45$\pm$2.57 & 73.9$\pm$3.87 & 75.32$\pm$2.9 & 74.74$\pm$3.63 & 74.97$\pm$2.87 & 74.37$\pm$3.46 \\
 & 10 & 78.68$\pm$1.74 & 77.16$\pm$2.55 & 78.26$\pm$2.09 & 77.55$\pm$2.29 & 77.57$\pm$2.12 & 76.94$\pm$2.83 \\
 & 20 & 80.13$\pm$1.97 & 79.66$\pm$1.64 & 79.99$\pm$2.15 & 80$\pm$2.22 & 80.09$\pm$1.98 & 80.08$\pm$1.99 \\
\multirow{4}{*}{th\_pud} & 0 & 42.51 & 42.71 & 43.76 & 43.3 & 46.81 & 43.07 \\
 & 5 & 58.05$\pm$2.53 & 59.83$\pm$2.35 & 60.02$\pm$2.62 & 61.18$\pm$2.74 & 61.12$\pm$2.95 & 60.15$\pm$2.05 \\
 & 10 & 61.71$\pm$2.17 & 63.57$\pm$1.72 & 63.85$\pm$1.9 & 65.14$\pm$1.67 & 65.4$\pm$1.87 & 63.34$\pm$1.75 \\
 & 20 & 65.05$\pm$1.28 & 66.39$\pm$1.38 & 66.62$\pm$1.08 & 67.99$\pm$1.41 & 68.72$\pm$1.28 & 66.27$\pm$1.36 \\
\multirow{4}{*}{tl\_trg} & 0 & 76.9 & 77.43 & 77.59 & 85.12 & 82.27 & 80.62 \\
 & 5 & 83.01$\pm$3.52 & 82.95$\pm$3.66 & 84.09$\pm$4.75 & 84.5$\pm$4.14 & 84.4$\pm$4.01 & 84.01$\pm$4.64 \\
 & 10 & 85.78$\pm$1.66 & 85.4$\pm$2.06 & 86.86$\pm$2.3 & 87.27$\pm$2.62 & 87.23$\pm$2.87 & 87.42$\pm$2.12 \\
 & 20 & 87.27$\pm$2.04 & 87.48$\pm$2.32 & 88.69$\pm$1.96 & 89.1$\pm$2.34 & 89.2$\pm$1.85 & 89.24$\pm$1.86 \\
\multirow{4}{*}{tl\_ugnayan} & 0 & 60.37 & 64.38 & 63.58 & 63.01 & 64.41 & 64.76 \\
 & 5 & 74.8$\pm$1.86 & 76.35$\pm$2.27 & 75.73$\pm$2.01 & 75.2$\pm$2.37 & 78.13$\pm$2.01 & 76.91$\pm$2.44 \\
 & 10 & 77.02$\pm$3.68 & 79.31$\pm$1.48 & 78.35$\pm$1.64 & 78.93$\pm$1.3 & 80.69$\pm$1.71 & 79.28$\pm$1.62 \\
 & 20 & 78.91$\pm$1.44 & 82$\pm$1.07 & 80.86$\pm$1.04 & 81.14$\pm$1.32 & 82.66$\pm$1 & 81.71$\pm$1.31 \\
\multirow{4}{*}{wbp\_ufal} & 0 & 26.64 & 24.55 & 28.62 & 27.21 & 27.96 & 30.18 \\
 & 5 & 58$\pm$4.23 & 56.83$\pm$4.94 & 57.07$\pm$4.67 & 58.52$\pm$4.98 & 59.13$\pm$4.86 & 59.68$\pm$6.1 \\
 & 10 & 64.72$\pm$1.72 & 63.34$\pm$4.41 & 64.51$\pm$3.43 & 65.94$\pm$3.88 & 65.2$\pm$4.03 & 66.32$\pm$3.63 \\
 & 20 & 71.84$\pm$3.39 & 66.67$\pm$3.67 & 70.45$\pm$3.46 & 70.6$\pm$3.29 & 67.98$\pm$3.77 & 70.75$\pm$3.55 \\
\multirow{4}{*}{wo\_wtb} & 0 & 34.79 & 33.05 & 34.72 & 34.11 & 34.09 & 35.27 \\
 & 5 & 46.12$\pm$2.41 & 45.47$\pm$2.7 & 45.86$\pm$2.36 & 46.69$\pm$2.23 & 47.49$\pm$2.66 & 46.49$\pm$2.3 \\
 & 10 & 50.01$\pm$2.03 & 48.49$\pm$1.69 & 49.13$\pm$2.18 & 49.69$\pm$1.79 & 50.97$\pm$2.1 & 49.67$\pm$2.1 \\
 & 20 & 53.32$\pm$1.19 & 51.27$\pm$1.39 & 52.73$\pm$1.65 & 52.79$\pm$1.15 & 53.97$\pm$1.45 & 52.58$\pm$1.55 \\
\multirow{4}{*}{yo\_ytb} & 0 & 41.46 & 47.34 & 45.31 & 45.59 & 50.45 & 49.1 \\
 & 5 & 59.59$\pm$3.02 & 62.93$\pm$2.71 & 61.66$\pm$2.54 & 61.26$\pm$2.8 & 64.5$\pm$2.51 & 63.3$\pm$3.09 \\
 & 10 & 63.34$\pm$ & 66.71$\pm$1.63 & 65.68$\pm$2 & 65.39$\pm$2.17 & 68.18$\pm$1.63 & 67.31$\pm$1.5 \\
 & 20 & 67.23$\pm$1.06 & 69.14$\pm$1.19 & 69.45$\pm$1.01 & 68.56$\pm$1.43 & 70.9$\pm$1.1 & 69.58$\pm$1.25 \\
\bottomrule
\end{tabular}
\caption{POS tagging results on all evaluation languages.}
\label{tab:res_ud2}
\end{table*}

\begin{figure*}[t]
    \centering
    \includegraphics[width=\textwidth]{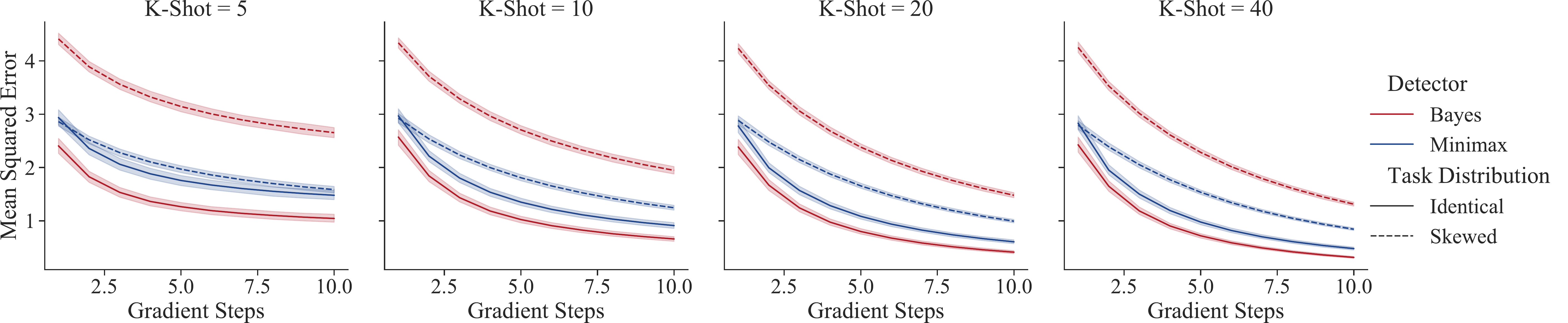}
    \caption{Mean Squared Error of MAML across gradient steps (from 1 to 10) of different criteria (B and MM) under identical and skewed task distributions. Each frame represents a separate run of fast adaptation with different amounts of target examples available ($k$-shot).}
    \label{fig:sinusoidal}
\end{figure*}
\rahul{Neyman-Pearson curves missing in Figure 3.}

\section{Additional Experiments \& Results}
\label{app:extraexp}

\paragraph{Additional Results.}
Table \ref{tab:res_ud2} contains POS tagging F$_1$ scores of all languages, for all models, in both zero and few-shot settings. Table \ref{tab:res_qa2} shows the exact match and F$_1$ scores for QA. 

\paragraph{Sinusoidal Regression.}
After delving into real-world, large-scale NLP applications, we additionally illustrate the effect of the alternative criteria on other ML domains. We run a proof-of-concept experiment on a toy task where we can fully control the distribution of the training and evaluation data, viz.\ regression of a sinusoidal function.

For this task, we follow the same experimental setting and hyper-parameters of \citet{finn2017model}: combinations of amplitudes $a \in [0.1, 5]$ and phases $p \in [0, \pi]$ determine a set of tasks characterised by the function $y = \sin(x- p) \cdot a$. The inputs are sampled at random from the interval $x \in [-5, 5]$.

While both train and evaluation tasks in the original version were sampled uniformly from \textit{identical} ranges, we also construct an alternative setting with \textit{skewed} distributions sampled from disjoint ranges: during training, $a \in [2.5, 5]$ and $p \in [\frac{\pi}{2}, \pi]$; during evaluation, $a \in [0.1, 2.5]$ and $p \in [0, \frac{\pi}{2}]$.

For Minimax MAML, we aim at learning the distribution over tasks adversarially. In particular, we consider two separate discrete categorical distributions for amplitudes $\softmax(\lv{\tau}_u^{(a)})$ and phases $\softmax(\lv{\tau}_u^{(p)})$ over their respective ranges discretised into 1,000 atoms. Hence, the probability of a task with the $i$-th amplitude value and the $j$-th phase value is simply $\tau_i^{(a)} \times \tau_j^{(p)}$. %

The results for sinusoidal regression are shown in Figure~\ref{fig:sinusoidal}. Vanilla MAML (Bayes criterion) consistently outperforms the minimax criterion when the task distribution is identical; on the other hand, the reverse occurs when the task distribution is skewed. MM performs much better in this case, with the gap in performance increasing as the shots $k$ decrease. This verifies our hypothesis that the minimax criterion should benefit out-of-distribution regression tasks.

\begin{table*}[p]
\centering
\small
\begin{tabular}{r|c|cccccc}
\toprule 
Dataset &  k &               J &               B &              MM &              NP &             MM+ &             NP+ \\
\midrule
\rowcolor{Gray}
\multicolumn{8}{c}{Exact Match} \\
\multirow{4}{*}{Arabic} &   0 &           48.97 &           49.29 &           51.47 &           51.36 &            49.4 &           48.64 \\
&          5 &   52.2$\pm$3.92 &  50.19$\pm$3.52 &  53.38$\pm$3.52 &   51.48$\pm$3.2 &  49.27$\pm$3.89 &  51.27$\pm$4.75 \\
&         10 &  54.51$\pm$2.47 &  52.81$\pm$2.93 &  54.96$\pm$2.93 &  53.67$\pm$2.16 &  52.05$\pm$3.27 &  53.67$\pm$3.43 \\
&         20 &     56$\pm$1.85 &  54.64$\pm$1.86 &  56.59$\pm$1.56 &  55.43$\pm$1.86 &  54.45$\pm$1.94 &  55.78$\pm$2.13 \\
\multirow{4}{*}{Bengali} &  0 &           45.13 &           46.02 &           51.33 &           44.25 &           45.13 &           51.33 \\
&          5 &  46.32$\pm$3.48 &   45.3$\pm$3.11 &  50.76$\pm$3.03 &   47.22$\pm$3.3 &     45$\pm$2.98 &  49.45$\pm$3.17 \\
&         10 &  47.22$\pm$3.15 &  46.44$\pm$3.08 &  50.83$\pm$2.94 &  49.39$\pm$3.84 &   45.98$\pm$3.7 &  50.01$\pm$3.22 \\
&         20 &  49.47$\pm$3.54 &  48.24$\pm$4.15 &  52.37$\pm$3.57 &  50.21$\pm$3.62 &  47.68$\pm$3.31 &  51.24$\pm$3.03 \\
\multirow{4}{*}{Finnish} &  0 &           42.33 &           43.61 &           47.95 &           49.36 &           47.83 &           46.42 \\
&          5 &   46.5$\pm$4.96 &  45.75$\pm$3.21 &  47.69$\pm$3.48 &  48.75$\pm$3.21 &  45.66$\pm$3.53 &  47.57$\pm$4.21 \\
&         10 &  48.56$\pm$2.65 &  47.25$\pm$2.81 &  49.43$\pm$2.78 &   50.28$\pm$3.1 &  46.85$\pm$2.77 &   48.55$\pm$3.1 \\
&         20 &  49.81$\pm$2.09 &  48.82$\pm$2.77 &  50.43$\pm$2.34 &  52.22$\pm$3.01 &  48.18$\pm$2.48 &  50.89$\pm$2.49 \\
\multirow{4}{*}{Korean} &   0 &              50 &           50.72 &           53.62 &           48.55 &           51.45 &           53.62 \\
&          5 &  51.37$\pm$2.52 &   49.5$\pm$2.76 &  51.87$\pm$2.11 &  49.52$\pm$2.48 &  49.57$\pm$2.35 &     52.17$\pm$2 \\
&         10 &  52.63$\pm$2.41 &  50.63$\pm$2.46 &  52.29$\pm$1.85 &   50.1$\pm$2.29 &  50.29$\pm$2.51 &     53$\pm$1.93 \\
&         20 &  54.07$\pm$2.11 &  51.88$\pm$2.15 &  53.55$\pm$1.91 &  51.87$\pm$2.03 &  51.71$\pm$2.13 &  53.67$\pm$2.16 \\
\multirow{4}{*}{Indonesian} & 0 &         56.46 &           51.86 &           54.87 &           56.28 &           52.74 &           56.28 \\
&          5 & 57.99$\pm$2.94 &  55.49$\pm$3.18 &  56.04$\pm$2.99 &   57.61$\pm$2.7 &  55.53$\pm$3.82 &  55.39$\pm$2.67 \\
&         10 &  59.4$\pm$2.49 &  57.11$\pm$2.81 &  58.54$\pm$2.49 &  58.59$\pm$1.96 &  57.08$\pm$2.84 &  56.86$\pm$1.95 \\
&         20 & 60.99$\pm$2.09 &  58.99$\pm$2.51 &  60.76$\pm$2.21 &   59.9$\pm$1.69 &  59.11$\pm$2.07 &  57.95$\pm$1.94 \\
\multirow{4}{*}{Russian} &   0 &          44.21 &           43.23 &           41.01 &           40.39 &           41.01 &           37.44 \\
&          5 &  49.45$\pm$4.36 &  47.41$\pm$3.92 &  46.66$\pm$4.01 &  46.83$\pm$4.34 &   46.2$\pm$4.61 &  44.09$\pm$5.38 \\
&         10 &  51.84$\pm$3.04 &  49.66$\pm$2.83 &  48.72$\pm$3.56 &  48.81$\pm$3.79 &  47.97$\pm$4.43 &  47.66$\pm$4.05 \\
&         20 &   53.6$\pm$2.45 &  50.72$\pm$2.55 &   51.05$\pm$2.8 &   51.5$\pm$2.75 &  50.47$\pm$2.45 &  50.25$\pm$2.96 \\
\multirow{4}{*}{Swahili} &  0 &           43.49 &           45.29 &           41.88 &           41.48 &           45.69 &           45.29 \\
&          5 &  46.47$\pm$5.11 &  49.07$\pm$4.31 &   48.9$\pm$4.88 &   47.6$\pm$4.21 &   48.8$\pm$4.28 &   47.32$\pm$4.3 \\
&         10 &  50.06$\pm$4.13 &  51.37$\pm$3.45 &   51.1$\pm$3.83 &  50.37$\pm$3.72 &  49.79$\pm$3.59 &  49.96$\pm$3.88 \\
&         20 &  54.02$\pm$3.06 &  53.82$\pm$2.63 &  53.94$\pm$2.54 &  52.16$\pm$2.89 &  52.51$\pm$3.47 &  52.65$\pm$3.26 \\
\multirow{4}{*}{Telugu} &   0 &            43.5 &           42.75 &           44.54 &              42 &            41.7 &           45.14 \\
&          5 &  45.97$\pm$2.85 &  44.58$\pm$3.44 &  45.33$\pm$3.91 &  44.89$\pm$3.44 &  41.87$\pm$5.35 &  42.92$\pm$5.36 \\
&         10 &   48.11$\pm$3.4 &   46.64$\pm$3.1 &  47.59$\pm$2.95 &   46.4$\pm$2.69 &  45.32$\pm$4.23 &   46.3$\pm$3.51 \\
&         20 &   50.1$\pm$2.55 &  49.08$\pm$2.42 &  49.21$\pm$2.77 &   48.8$\pm$1.97 &  47.11$\pm$2.71 &  48.88$\pm$2.91 \\
\rowcolor{Gray}
\multicolumn{8}{c}{F$_1$ score} \\
\multirow{4}{*}{Arabic} &   0 &           65.57 &           67.38 &           66.59 &           67.44 &           64.98 &           65.45 \\
&          5 &   68.4$\pm$3.82 &  67.09$\pm$3.51 &  69.66$\pm$3.45 &  67.59$\pm$3.26 &  65.92$\pm$3.93 &  67.76$\pm$4.86 \\
&         10 &  70.56$\pm$2.47 &  69.55$\pm$2.88 &  71.35$\pm$2.83 &  69.82$\pm$2.15 &  68.82$\pm$3.64 &  70.28$\pm$3.63 \\
&         20 &  72.14$\pm$1.78 &  71.14$\pm$1.87 &  73.21$\pm$1.46 &  71.55$\pm$1.88 &   71.5$\pm$1.99 &  72.26$\pm$2.31 \\
\multirow{4}{*}{Bengali} &  0 &           57.24 &           62.57 &           66.29 &           59.64 &           60.28 &           62.86 \\
&          5 &  59.27$\pm$2.79 &  60.04$\pm$2.97 &  64.65$\pm$2.84 &   61.71$\pm$3.1 &  59.46$\pm$2.72 &  61.85$\pm$2.65 \\
&         10 &   59.88$\pm$2.7 &  60.64$\pm$2.86 &  64.88$\pm$2.71 &  63.52$\pm$3.38 &  60.03$\pm$3.28 &  62.33$\pm$2.89 \\
&         20 &  62.11$\pm$3.15 &   62.1$\pm$3.51 &  65.93$\pm$3.07 &  64.31$\pm$2.95 &  61.72$\pm$2.96 &  63.86$\pm$2.72 \\
\multirow{4}{*}{Finnish} &  0 &           61.85 &           63.57 &           61.72 &           63.79 &           62.12 &           61.64 \\
&          5 &  61.48$\pm$3.47 &  61.76$\pm$2.63 &  61.66$\pm$2.84 &   62.66$\pm$2.8 &  60.49$\pm$2.42 &  61.57$\pm$3.94 \\
&         10 &  62.98$\pm$1.53 &  62.48$\pm$2.03 &  63.26$\pm$2.31 &   64.14$\pm$2.7 &  61.58$\pm$2.15 &  62.58$\pm$2.91 \\
&         20 &  63.81$\pm$1.57 &  63.66$\pm$2.07 &   64.65$\pm$2.2 &  65.64$\pm$2.84 &     63$\pm$2.24 &   64.9$\pm$2.27 \\
\multirow{4}{*}{Korean} &   0 &           60.26 &           62.71 &            62.4 &           58.68 &            61.2 &           64.35 \\
&          5 &  61.31$\pm$2.47 &  60.82$\pm$2.47 &  61.67$\pm$2.17 &  59.31$\pm$2.34 &  59.27$\pm$2.33 &  62.13$\pm$2.01 \\
&         10 &  62.52$\pm$2.26 &   62.02$\pm$2.1 &  61.86$\pm$2.05 &  60.03$\pm$2.15 &  59.86$\pm$2.51 &  62.92$\pm$1.91 \\
&         20 &  64.04$\pm$2.01 &  63.08$\pm$1.87 &  63.43$\pm$1.89 &  61.84$\pm$1.97 &  61.52$\pm$1.89 &  63.55$\pm$1.95 \\
\multirow{4}{*}{Indonesian} & 0 &         69.96 &           65.99 &           70.05 &           70.82 &           68.02 &           70.19 \\
&          5 &   71.4$\pm$2.81 &  69.22$\pm$3.28 &  70.61$\pm$2.74 &  71.18$\pm$2.17 &   69.95$\pm$3.4 &  69.37$\pm$2.58 \\
&         10 &  72.69$\pm$2.27 &  70.79$\pm$2.78 &  72.79$\pm$2.53 &  72.23$\pm$1.77 &  71.33$\pm$2.46 &  70.93$\pm$1.96 \\
&         20 &  74.11$\pm$1.79 &  72.49$\pm$2.57 &  74.65$\pm$2.11 &  73.52$\pm$1.45 &   73.11$\pm$1.8 &  71.96$\pm$1.94 \\
\multirow{4}{*}{Russian} &  0 &           65.93 &           64.15 &           64.47 &            63.2 &           64.13 &           61.08 \\
&          5 &  66.96$\pm$1.52 &   65.11$\pm$1.5 &  65.03$\pm$1.39 &  65.13$\pm$1.33 &  64.58$\pm$1.45 &  62.17$\pm$1.61 \\
&         10 &  67.86$\pm$1.15 &  66.21$\pm$1.28 &  65.84$\pm$1.65 &  65.89$\pm$1.33 &  65.53$\pm$1.46 &  63.63$\pm$1.67 \\
&         20 &   68.7$\pm$1.01 &  66.85$\pm$1.38 &  66.94$\pm$1.45 &   67.1$\pm$1.38 &   66.4$\pm$1.19 &  65.01$\pm$1.82 \\
\multirow{4}{*}{Swahili} &  0 &           60.01 &           62.63 &           59.84 &           58.74 &           64.13 &           62.48 \\
&          5 &  60.21$\pm$4.38 &   62.7$\pm$3.37 &  62.48$\pm$3.72 &   61.9$\pm$3.41 &  63.43$\pm$3.38 &  61.81$\pm$4.06 \\
&         10 &  62.62$\pm$2.66 &  64.36$\pm$2.31 &  63.79$\pm$3.19 &   63.6$\pm$3.02 &  63.77$\pm$2.88 &  63.78$\pm$3.06 \\
&         20 &  65.18$\pm$2.09 &     65.89$\pm$2 &  66.27$\pm$1.99 &   65.48$\pm$1.7 &  66.21$\pm$2.19 &  66.12$\pm$2.11 \\
\multirow{4}{*}{Telugu} &   0 &           52.43 &            51.1 &            53.1 &           52.83 &           51.93 &           53.96 \\
&          5 &  60.99$\pm$5.15 &   59.6$\pm$6.05 &  59.21$\pm$6.17 &  61.27$\pm$5.05 &  57.91$\pm$6.64 &  57.21$\pm$7.33 \\
&         10 &  63.99$\pm$3.92 &  62.92$\pm$4.19 &  62.85$\pm$3.62 &  62.63$\pm$4.98 &   61.92$\pm$5.1 &  61.74$\pm$5.18 \\
&         20 &  65.96$\pm$2.37 &  65.29$\pm$2.15 &  64.53$\pm$3.06 &  65.63$\pm$1.61 &  63.67$\pm$2.58 &   64.9$\pm$3.26 \\
\bottomrule
\end{tabular}
\caption{QA results on all evaluation languages.}
\label{tab:res_qa2}
\end{table*}

\end{document}